\documentclass{evolang}
\usepackage{graphicx}
\usepackage{url}
\usepackage{authblk}

\usepackage{comment}
\usepackage{multirow}

\begin{document}

\title{Construction and Evaluation of a Self-Attention Model for Semantic Understanding of Sentence-Final Particles}
\author[*1]{Shuhei Mandokoro}
\author[2]{Natsuki Oka}
\author[1]{Akane Matsushima}
\author[1]{Chie Fukada}
\author[3]{Yuko Yoshimura}
\author[4]{Koji Kawahara}
\author[1]{Kazuaki Tanaka}
\affil[*]{Corresponding Author: m2622033@edu.kit.ac.jp}
\affil[1]{Kyoto Institute of Technology, Kyoto, Japan}
\affil[2]{Miyazaki Sangyo-keiei University, Miyazaki, Japan}
\affil[3]{Kanazawa University, Kanazawa, Japan}
\affil[4]{Nagoya University of Foreign Studies, Nagoya, Japan}



\maketitle

Sentence-final particles in Japanese play critical role in expressing the speaker’s mental attitudes. They are acquired at an early age and occur frequently in everyday conversation. The computational model of language acquisition is gaining popularity. However, few computer models of the acquisition of sentence-final particles have been proposed
\BBOP\shortciteNP{Oka13}\BBC Matsushima, Kanejiri \BOthers,  \citeyearNP{Matsushima19a}\BBC Matsushima, Oka \BOthers,  \citeyearNP{Matsushima19b}\BBCP.

In recent years, models using self-attention \shortcite{attention17,BERT} have become mainstream in natural language processing because of their high performance. Thereafter, models that process inputs from diverse modalities, in addition to the language, have been proposed \shortcite{Lu19,Chen19,Radford21,Perceiver,H-Perceiver}. Following this trend, 
we propose Subjective BERT, in which the input is extended to language and five subjective senses: vision, inference, taste, hunger, and desire, thus enabling a deeper understanding of words.

The simulation experiment was conducted as follows: Assuming a scenario in which a caregiver addresses a child, the utterances and situation were fed into Subjective BERT as tokens. Pre-training was performed using masked language modeling \cite{BERT}. No fine-tuning was performed. We prepared 470 data chunks: 440 for pre-training, and 30 for the test trial. 
For every 440 data chunks, 50 masked training examples were generated, resulting in 22,000 examples,which were given to the Subjective BERT as one epoch.
The following is the list of tokens:
\begin{itemize}
    \item Previous and current utterances: no utterance or one of the ten variations of utterance, including {\it Ringo-da-yo} (``Look at the apple which I am looking at.''), {\it Oishi-souda-yo} (``I want to tell you that it looks tasty.''), {\it Onakasuita-ne} (``I want to make sure that we both feel hungry.''), 
    and  {\it Tabetai-ne} (``I want to make sure that we both want to eat.''). Hyphens in Japanese indicate word separators when the utterances are input into Subjective BERT.
    \item Current vision: A delicious-looking \{apple/banana\} or a \{green apple/spotted banana\} that the child does not find appealing.
    \item Current inference of taste: One of the four inferences of taste, all of which are assumed to be inferred from the child's current visions.
    \item Current taste: one of three options: delicious apple taste, delicious banana taste, or none (when the child is not eating).
    \item Current hunger: hungry or not hungry.
    \item Current desire: no desire or desire to eat \{an apple/a banana\}.
\end{itemize}

\begin{figure}[b]
\begin{center}
\scalebox{0.421}{\includegraphics{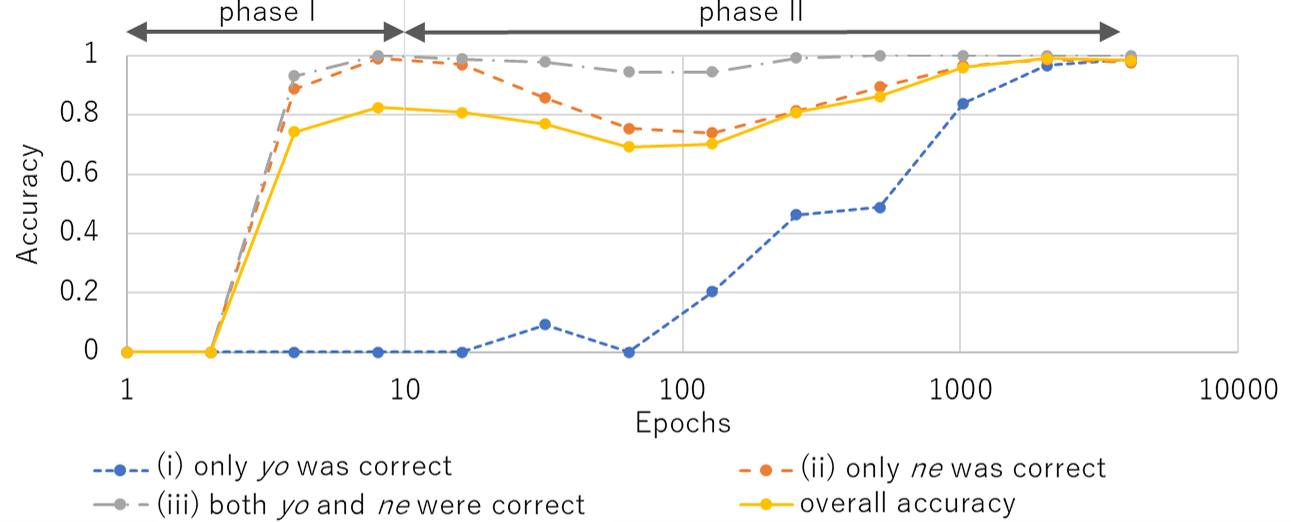}}
\end{center}
\caption{{\footnotesize U-shaped learning progression of sentence-final particles.}}
\label{learning progression}
\end{figure}

At several points during pre-training, we assessed the learning progression of Subjective BERT using the test dataset to determine whether it can predict a correct sentence-final particle in each masked token. Each of the thirty interaction data in the test set contained two utterances, in which at most two sentence-final particles were included. Three types of questions were included in the fill-in test: (i) only {\it yo} was correct; (ii) only {\it ne} was correct; and (iii) both {\it yo} and {\it ne} were correct. The average ratio of the three question types was approximately 1:3:2. The holdout method was repeated six times while re-selecting  training and test sets. Figure~\ref{learning progression} shows the transition of the correct response rates averaged from the data of the six trials.

Two phases were observed in the learning process of the two sentence-final particles. In phase~I, the accuracy of type (ii) and type (iii) questions increased rapidly and reached almost 100\% at around ten epochs. The accuracy of type (i) questions, on the other hand, remained 0\%. In the 8th epoch, the learning system answered {\it ne} to almost all three types of fill-in-the-blank questions. The difference in the prediction accuracy for {\it ne} and {\it yo} can be attributed to the fact that {\it ne} occurred nearly twice as often as {\it yo} in the input data of the current study. This frequency bias was caused by the exclusion of unnatural speech as infant-directed speech\footnote{The utterances on Current hunger and Current desire appeared only with {\it ne}.} from the dataset. The investigation of the frequency distribution in actual infant-directed speech is a topic for future work.

In phase~II, the accuracy of {\it ne}  fell slightly below 100\% and then it gradually increased again to 100\% after approximately 2000 epochs. Slowly, the accuracy of {\it yo} began to increase for the first time in phase~II and reached 100\% after approximately 2000 epochs. We analyzed the trend of wrong answers during the low accuracy of type (ii) questions and found that {\it ne} was often mistaken for the plain copula {\it da} in the 32nd and 64th epochs. We also found that as the learning of {\it yo} progressed, the percentage of {\it yo} being wrongly proffered as the answer to the questions where {\it ne} was correct increased from Epoch 128 onward.

In all three cases, (i), (ii), and (iii), after some fluctuations, the final correct response rates reached approximately 100\%. 
This indicates that the model could eventually correctly predict the sentence-final particles {\it yo} and {\it ne}. Since we assume that the function of the two Japanese sentence-final particles is to associate input utterances with the other input information, i.e., the five pieces of the sense modality information, and that to understand the meanings of these particles is to understand this kind of associations, we can say that the model acquired the meanings of the sentence-final particles {\it yo} and {\it ne} in the present study. What the model learned were:

\begin{enumerate}
    \item When the current utterance contained {\it ne}, the model associated the utterance with the simultaneously input sense modality information which matched the content word in the utterance. For example, when the caregiver said {\it Oishii-ne} (``I want to make sure that we are experiencing a delicious taste sensation.''), the system associated the utterance with a sense modality token concerning ``delicious taste'' which was simultaneously input into the model.
    \item In contrast, when the utterance contained {\it yo}, the model associated the utterance with the sense modality information which were relevant to the content word in the utterance in the next period. For example, when the caregiver said {\it Ringo-da-yo} (``Look at the apple which I am looking at.''), the system directed its attention to the apple at the next moment and associated the utterance with a visual image of the apple the system obtained.
\end{enumerate}

Currently, we are expanding the variation of the language input into Subjective BERT. Next, we will investigate the processes that occur in the learning system more closely, particularly focusing on  how self-attention
changes as the learning phase progresses.

\section*{Acknowledgements}
This work was supported by JSPS KAKENHI Grant Number JP20H05004.

\bibliographystyle{apacite}
\bibliography{main} 

\begin{thebibliography}{}

\bibitem[\protect\BCAY{Carreira \BOthers{}}{Carreira
  \BOthers{}}{2022}]{H-Perceiver}
Carreira, J., Koppula, S., Zoran, D., Recasens, A., Ionescu, C., Henaff, O.,
  Shelhamer, E., Arandjelovic, R., Botvinick, M., Vinyals, O., Simonyan, K.,
  Zisserman, A.\BCBL{} \BBA{} Jaegle, A.
\newblock{}\BBOP{}2022\BBCP{}.
\newblock{}\BBOQ{}Hierarchical perceiver.\BBCQ{}
\newblock{}\Bem{arXiv preprint arXiv:2202.10890}.

\bibitem[\protect\BCAY{Chen \BOthers{}}{Chen \BOthers{}}{2019}]{Chen19}
Chen, Y.-C., Li, L., Yu, L., Kholy, A.~E., Ahmed, F., Gan, Z., Cheng, Y.\BCBL{}
  \BBA{} Liu, J.
\newblock{}\BBOP{}2019\BBCP{}.
\newblock{}\BBOQ{}Uniter: Universal image-text representation learning.\BBCQ{}
\newblock{}\Bem{arXiv preprint arXiv:1909.11740}.

\bibitem[\protect\BCAY{Devlin, Chang, Lee\BCBL{} \BBA{} Toutanova}{Devlin
  \BOthers{}}{2019}]{BERT}
Devlin, J., Chang, M.-W., Lee, K.\BCBL{} \BBA{} Toutanova, K.
\newblock{}\BBOP{}2019\BBCP{}.
\newblock{}\BBOQ{}{BERT}: Pre-training of deep bidirectional transformers for
  language understanding.\BBCQ{}
\newblock{}In \Bem{Proceedings of the 2019 conference of the north {A}merican
  chapter of the association for computational linguistics: Human language
  technologies, volume 1 (long and short papers)}\ (\BPGS\ 4171--4186).
\newblock{}Minneapolis, Minnesota: Association for Computational Linguistics.

\bibitem[\protect\BCAY{Jaegle \BOthers{}}{Jaegle \BOthers{}}{2021}]{Perceiver}
Jaegle, A., Gimeno, F., Brock, A., Zisserman, A., Vinyals, O.\BCBL{} \BBA{}
  Carreira, J.
\newblock{}\BBOP{}2021\BBCP{}.
\newblock{}\BBOQ{}Perceiver: General perception with iterative
  attention.\BBCQ{}
\newblock{}\Bem{arXiv preprint arXiv:2103.03206}.

\bibitem[\protect\BCAY{Lu, Batra, Parikh\BCBL{} \BBA{} Lee}{Lu
  \BOthers{}}{2019}]{Lu19}
Lu, J., Batra, D., Parikh, D.\BCBL{} \BBA{} Lee, S.
\newblock{}\BBOP{}2019\BBCP{}.
\newblock{}\BBOQ{}Vilbert: Pretraining task-agnostic visiolinguistic
  representations for vision-and-language tasks.\BBCQ{}
\newblock{}\Bem{arXiv preprint arXiv:1908.02265}.

\bibitem[\protect\BCAY{Matsushima, Kanajiri, Hattori, Fukada\BCBL{} \BBA{}
  Oka}{Matsushima \BOthers{}}{2019}]{Matsushima19a}
Matsushima, A., Kanajiri, R., Hattori, Y., Fukada, C.\BCBL{} \BBA{} Oka, N.
\newblock{}\BBOP{}2019\BBCP{}.
\newblock{}\BBOQ{}Stepwise acquisition of dialogue act through human-robot
  interaction.\BBCQ{}
\newblock{}In \Bem{Conference papers of the {International} {Joint}
  {Conference} on {Neural} {Networks}.}
\newblock{}

\bibitem[\protect\BCAY{Matsushima, Oka, Fukada\BCBL{} \BBA{} Tanaka}{Matsushima
  \BOthers{}}{2019}]{Matsushima19b}
Matsushima, A., Oka, N., Fukada, C.\BCBL{} \BBA{} Tanaka, K.
\newblock{}\BBOP{}2019\BBCP{}.
\newblock{}\BBOQ{}Understanding dialogue acts by bayesian inference and
  reinforcement learning.\BBCQ{}
\newblock{}In \Bem{Proceedings of the 7th international conference on
  human-agent interaction}\ (\BPG\ 262–264).
\newblock{}New York, NY, USA: Association for Computing Machinery.

\bibitem[\protect\BCAY{Oka, Wu, Fukada\BCBL{} \BBA{} Ozeki}{Oka
  \BOthers{}}{2013}]{Oka13}
Oka, N., Wu, X., Fukada, C.\BCBL{} \BBA{} Ozeki, M.
\newblock{}\BBOP{}2013\BBCP{}.
\newblock{}\BBOQ{}Concurrent acquisition of the meaning of sentence-final
  particles and nouns through human-robot interaction.\BBCQ{}
\newblock{}In M.~Lee, A.~Hirose, Z.-G. Hou\BCBL{} \BBA{} R.~M. Kil\ (\BEDS),
  \Bem{{ICONIP} 2013, {Part I}, {LNCS} 8226}\ (\BPGS\ 387--394).
\newblock{}Heidelberg: Springer.

\bibitem[\protect\BCAY{Radford \BOthers{}}{Radford
  \BOthers{}}{2021}]{Radford21}
Radford, A., Kim, J.~W., Hallacy, C., Ramesh, A., Goh, G., Agarwal, S., Sastry,
  G., Askell, A., Mishkin, P., Clark, J., Krueger, G.\BCBL{} \BBA{} Sutskever,
  I.
\newblock{}\BBOP{}2021\BBCP{}.
\newblock{}\BBOQ{}Learning transferable visual models from natural language
  supervision.\BBCQ{}
\newblock{}\Bem{arXiv preprint arXiv:2103.00020}.

\bibitem[\protect\BCAY{Vaswani \BOthers{}}{Vaswani
  \BOthers{}}{2017}]{attention17}
Vaswani, A., Shazeer, N., Parmar, N., Uszkoreit, J., Jones, L., Gomez, A.~N.,
  Kaiser, L.\BCBL{} \BBA{} Polosukhin, I.
\newblock{}\BBOP{}2017\BBCP{}.
\newblock{}\BBOQ{}Attention is all you need.\BBCQ{}
\newblock{}\Bem{arXiv preprint arXiv:1908.02265}.

\end{thebibliography}

\end{document}